\documentclass[10pt,twocolumn,letterpaper]{article}
\usepackage{float} 
\usepackage{cvpr}
\usepackage{times}
\usepackage{epsfig}
\usepackage{graphicx}
\usepackage{amsmath}
\usepackage{amssymb}
\usepackage{multirow}
\usepackage{enumitem}
\usepackage{url} 
\usepackage[marginal]{footmisc}

\usepackage[breaklinks=true,bookmarks=false]{hyperref}

\cvprfinalcopy 

\def\cvprPaperID{****} 

\begin{document}

\title{Vehicle Re-Identification Based on Complementary Features}

\author{Cunyuan Gao\textsuperscript{1}, Yi Hu, Yi Zhang, Rui Yao\thanks{corresponding author. Email: ruiyao@cumt.edu.cn} \textsuperscript{1}, Yong Zhou\textsuperscript{1}, Jiaqi Zhao\textsuperscript{1} \\
\textsuperscript{1}School of Computer Science and Technology, China University of Mining and Technology, China
\\
{
{\tt\small cunyuangao@cumt.edu.cn, huyibupt@gmail.com, eezhangyi@zju.edu.cn, ruiyao@cumt.edu.cn}}
}

\maketitle

\begin{abstract}
In this work, we present our solution to the vehicle re-identification (vehicle Re-ID) track in AI City Challenge 2020 (AIC2020). The purpose of vehicle Re-ID is to retrieve the same vehicle appeared across multiple cameras, and it could make a great contribution to the Intelligent Traffic System(ITS) and smart city. Due to the vehicle's orientation, lighting and inter-class similarity, it is difficult to achieve robust and discriminative representation feature. For the vehicle Re-ID track in AIC2020, our method is to fuse features extracted from different networks in order to take advantages of these networks and achieve complementary features. For each single model, several methods such as multi-loss, filter grafting, semi-supervised are used to increase the representation ability as better as possible. Top performance in City-Scale Multi-Camera Vehicle Re-Identification demonstrated the advantage of our methods, we got 5-th place in the vehicle Re-ID track of AIC2020. The codes are available at \url{https://github.com/gggcy/AIC2020_ReID}.
\end{abstract}


\section{Introduction}

With the development of ITS, vehicle Re-ID has attracted increasing attention in computer vision community \cite{zhou2018aware, zhu2019vehicle, lou2019embedding}. The target of vehicle Re-ID is to find the vehicles in the gallery that have the same identical as the query vehicle. Compared with other image retrieval tasks, vehicle Re-ID is more challenging because of two main reasons. Firstly, same identical  vehicles may have different orientations in different cameras and the appearance of the front part and the rear part of a vehicle is much dissimilar. Secondly, two different vehicles with same brand and serial have completely same appearance except of very trivial difference. 

In the past, license plate recognition was a conventional and efficient vehicle Re-ID solution. However, the license plate of the vehicle may be blocked, dirt, or the video is not clear enough to be clearly seen. Also, there are confusing letters making license plate recognition unreliable. Therefore, vehicle Re-ID through vehicle visual feature is an essential part.

At present, feature extraction and metric learning are the mainstream research directions in the field of vehicle Re-ID. Compared with traditional manual selection features, Convolutional Neural Networks (CNN) can automatically learn discriminative high-level semantic features according to task requirements, which greatly improves the performance of vehicle Re-ID. Due to occlusion, lighting and diverse viewpoints, there are still many challenging problems.

In recent years, a few powerful networks \cite{he2016deep, huang2017densely} have been proposed. These networks utilize residual block or densely connected block to increase the network's depth and make the deep networks extract more discriminative features from the training images. Specially, for vehicle Re-ID task, part-based method \cite{he2019part} aims to combine the local part feature and global feature to increase the performance.
In this paper, our solution to the vehicle Re-ID track in AIC2020 could be summarized as follows:
\begin{itemize}
    \item We train multiple CNNs with a few methods including multi-loss, filter grafting, part-based feature learning, etc. These methods could do much more help to increase each single network's representation ability.
    \item We use semi-supervised method to annotate images in the test set with a fake label, and combine these fake-labeled images with the original training set to improve the final performance.
    \item Several post-processing methods are utilized to further make a progress of the result.
\end{itemize}

\begin{figure}[htbp]
\centering
\includegraphics[width=3.3in]{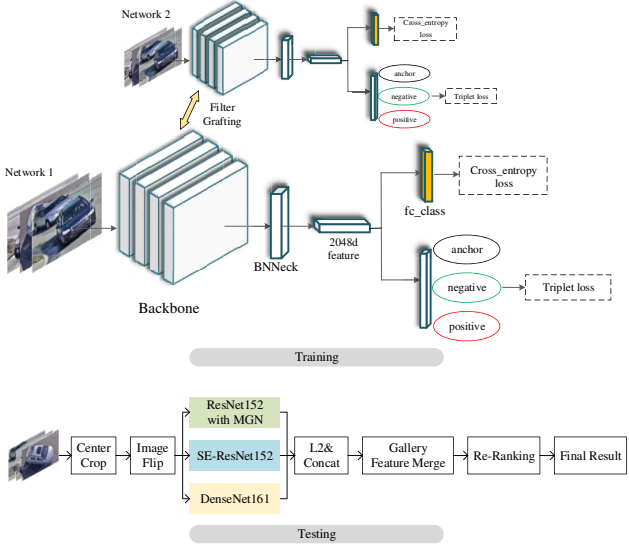}
\caption{Overview of our basic framework.}
\label{fig_1}

\end{figure}

\section{Related Works}
\subsection{Vehicle Re-identificaiton}
As several vehicle datasets \cite{liu2016large, bai2018group, zapletal2016vehicle} have been proposed, vehicle Re-ID has attracted increasing attention in recent several years. The current work of vehicle Re-ID is still focused on the feature level, and how to extract more informative features. While vehicle RE-ID is also improved significantly in recent years because of the development of CNN. Zhu \emph{et al.} \cite{zhu2019vehicle} and Guo \emph{et al.} \cite{guo2018learning} seek a better feature encoding method. Another solution is based on the part to extract highly distinguished features \cite{zhang2014part,wei2016mask,he2019part,liu2018ram}, these methods mainly rely on strong supervision information of each key part. However, the cost of annotation is high and the practical application is limited. Zhou \emph{et al.} \cite{zhou2018aware} adopt a viewpoint-aware attention model and a adversarial training architecture to implement effective multi-view
feature inference from single-view input.
Zhou \emph{et al.} \cite{zhou2018vehicle}
focus on the uncertainty of vehicle viewpoint in Re-ID, and propose an Adversarial Bi-directional LSTM Network (ABLN). Inspired by the behavior in human's recognition process, Chu \emph{et al.} \cite{chu2019vehicle} propose a novel viewpoint-aware metric learning approach, which learns two metrics for similar viewpoints and different viewpoints in two feature space.

\subsection{Re-ranking}
When vehicle Re-ID is regarded as a retrieval process, re-ranking is a key step to improve its performance. Re-ranking is mainly studied in generic instance retrieval \cite{chum2007total,qin2011hello,jegou2007contextual,shen2012object}. The main advantage of many re-ranking  methods is that it can be implemented without the need for additional training samples and can be applied to any initial ranking results.

A popular re-ranking approach is k-reciprocal encoding \cite{zhong2017re}. By encoding k-reciprocal nearest neighbors into a single vector. In order to obtain similar relationships from similar samples, a local expansion query is proposed to obtain more robust k-reciprocal features. The final distance based on the combination of the original distance and the Jaccard distance effectively improves the Re-ID performance on multiple large-scale data sets.

\section{Our Approach}
\subsection{Basic framework}
The basic framework of our approach is shown in the Fig. \ref{fig_1}. During training phases, the training images with ID label are sent into the backbone and generate a 2048-d feature and its predicted ID label. Two networks with the same structure will be trained in parallel and perform filter grafting. Then we apply two different loss functions as the optimization objective. The first one is hard triplet loss \cite{hermans2017defense}, it will enlarge the feature's distance between two different label samples and reduce the distance between samples with same label. The triplet loss can be defined as:
\begin{small}
$$L_{\text {triplet}}=\frac{1}{N}\sum_{i}^{N}\left[\left\|f\left(x_{i}^{a}\right)-f\left(x_{i}^{p}\right)\right\|_{2}^{2}-\left\|f\left(x_{i}^{a}\right)-f\left(x_{i}^{n}\right)\right\|_{2}^{2}+\alpha\right]_{+},$$
\end{small}
where $f\left(x_{i}^{a}\right)$, $f\left(x_{i}^{p}\right)$, $f\left(x_{i}^{n}\right)$ are the features extracted from anchor, positive and negative samples receptively and $N$ is the total number of training images in one batch. In theory, in order to ensure the best effect of network training, we must choose hard positive and hard negative,
$$\operatorname{argmax}_{x_{i}^{p}}\left\|f\left(x_{i}^{a}\right)-f\left(x_{i}^{p}\right)\right\|_{2}^{2},$$
$$\operatorname{argmin}_{x_{i}^{n}}\left\|f\left(x_{i}^{a}\right)-f\left(x_{i}^{n}\right)\right\|_{2}^{2}.$$

The second one is cross entropy loss. Given a batch of training images, we denote $y_i$ as the ground truth ID label, $p_{ij}$ as ID prediction logits of the $i$-th image on $j$-th class and $N$ as the total number of traing images in one batch. The cross entropy loss can be defined as:
$$L_{\text {cross}}=-\frac{1}{N}\sum_{i=1}^{N}\log \left(p_{ij}\right)$$ 
where $j$=$y_i$. Finally, we combine these two losses with a balanced weight, which is defined as:
$$L=L_{cross}+\alpha L_{hard triplets},$$
where $\alpha$ is the balanced weight of the triplet loss.

BNNeck \cite{luo2019bag} refers to the addition of a batch normalization (BN) layer after backbone.

On the basis of the framework, we add more useful modules to improve the single model's performance. One is MGN \cite{wang2018learning}, this module splits the input sample horizontally into several parts and  calculates each part's loss individually, while the global feature is also taken into loss calculation. The second one is self-attention constrain (SAC) \cite{jiang2018self} which makes the network pay more attention to some subtleties. Another one is squeeze-and-excitation(SE) block \cite{hu2018squeeze}, it is also an attention module that could enhance the feature's representation power by performing dynamic channel-wise feature re-calibration. 

During the testing phase, the trained network model will extract 2048-d feature from each input image. Features from different networks are concatenated together for vehicle Re-ID.

\subsection{Filter grafting}
\begin{figure}[htbp]
\centering
\includegraphics[width=3.3in]{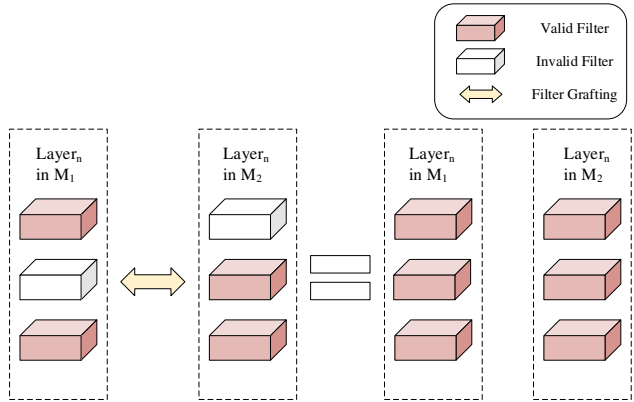}
\caption{Filter grafting with two networks. The two networks $M_{1}$ and $M_{2}$ accept filter parameters information from each other.}
\label{fig_grafting}

\end{figure}

Filter grafting \cite{meng2020filter} is a learning paradigm which aims to reactivate invalid filters during training such that the representation power could be improved. The weight of filters that are valid for other networks is grafted to the filters that are not valid for the self-network. Multiple networks are grafted together to promote progress The main framework is shown in Fig. \ref{fig_grafting}, given a network, two training processes are started parallelly
denoted as $M_1$ and $M_2$, respectively. In each epoch training, we graft the effective filter weight of $M_1$ into the invalid filter of $M_2$. The grafting in this process occurs at the convolution level, not the filter level, which means that we graft all filter weights in a certain layer in $M_1$ to $M_2$ in the same layer. Conversely, the same principle is used for grafting filter $M_2$ into $M_1$.  Filter grafting does not change the network structure. When performing grafting, the specific operation is shown in the following formula:
$$W_{i}^{M_{2}^{\prime}}=\alpha W_{i}^{M_{2}}+(1-\alpha) W_{i}^{M_{1}} \quad(0<\alpha<1).$$

After weighting the filter parameters of $M_1$ and $M_2$, the final filter parameters after grafting are obtained. We use the following formula to constrain $\alpha$ parameters
$$\alpha=A *\left(\arctan \left(C *\left(H\left(W_{i}^{M_{2}}\right)-H\left(W_{i}^{M_{1}}\right)\right)\right)\right)+0.5,$$
where $A$ and $C$ are two hyper-parameters and $H(*)$ is the entropy of the variables.

Although the two networks are trained in parallel when training, only use one of the grafted networks for testing.

\subsection{Semi-supervised learning}
The annotations of test set images are not accessible, so we propose a semi-supervised learning method to annotate test images with a fake label. Therefore, the test set becomes a supplementary of the training set. The steps of this method is as follows:
\begin{itemize}
    \item [1)] Train Vehicle Re-ID models using the original training set.
    \item [2)] Extract feature from the images in the test set using the trained Re-ID models.
    \item [3)] Cluster the testing images based on their features, and assign each cluster a fake label as the ID label.
    \item [4)] Merge the fake-labeled test set and the original training set to be a new training set. Then we train Re-ID models using the new training set.
\end{itemize}  
The number of IDs in the test set is 333. Therefore, the clustering algorithm we use is the k-means algorithm, and the clustering center is set to 333.

\subsection{Post-processing}
For the training results of multiple single models, we utilize several post-processing techniques.

\textbf{Center crop. }For the test set query and gallery, there is generally a bit of background redundancy. When we extract the test set features, we use center crop to make the image more focused on the vehicle itself and reduce the impact of the surrounding background.

\textbf{Image flipping. }When the model is inferred forward, we input the vehicle image and the horizontal flip of the vehicle image, and average the two feature values as the final feature of the vehicle. This can mitigate the impact of angle on vehicle feature expression.

\textbf{Model ensemble. }Model ensemble is proven to be effective in detection, classification, and Re-ID tasks. Different tasks have different ensemble methods. For Re-ID task, we have tried a lot of ensemble methods, and find out that it is better to expand the feature dimensions by concatenating features than to calculate the average feature of different models. Given the above trained networks, we could extract feature vectors for an image $x_{i}$, the feature vector can be expressed as $f_{i}^{j}$ of the $j-$th model. $f_{i}=\left[\left|f_{i}^{1}\right|,\left|f_{i}^{2}\right|, \ldots\left|f_{i}^{n}\right|\right]$. $|\cdot|$ represents $L2$ Normalization, and then we concatenate them as the final feature representations for an image $x_{i}$. The vehicle features extracted by each model are concatenated, and in our subsequent experiments, it is proved that the performance of the model would be improved compared to the single model.

\textbf{Query expansion. } Query expansion is also a re-ranking method. Through query expansion, the retrieval recall rate can be improved. Especially when the accuracy of top $K$ is high, the effect of this method will be significantly improved. After the first retrieval, the feature of query and features of top $K$ images retrieved in the gallery are averaged as the feature of the query, and then the second retrieval is performed, and the above operation is repeated $num$ times. The parameters $K$ and $num$ need to be adjusted for the official test set.

\textbf{Gallery feature merge. }The official test set also provides track id information, which contains gallery's 798 vehicle tracks, and each track is multiple pictures of the same vehicle. This is a prior information that can be used. The post-processing method for this prior information is to average the $T$ picture features of each track as the features of the $T$ pictures under this track. There is also a parameter $T$ that needs to be adjusted, or $T$ is all the pictures under this track. This post-processing method can be applied to some Re-ID tasks that provide video track information.

\textbf{Re-ranking with k-reciprocal encoding. }
K-reciprocal encoding \cite{zhong2017re} is an effective re-ranking method. A basic assumption is that if the returned image is sorted within the $k$ nearest neighbors of probe, then it is likely to be a true match. The idea of this method is to use query-to-test, query-to-query and test-to-test correlation in order to take the query and test distribution into consideration when ranking the results.

\section{Experiments}
\subsection{Datasets analyse}
CityFlow \cite{tang2019cityflow} is one of the largest vehicle Re-ID datasets. The dataset contains 56,277 bounding boxes of 666 vehicle identities, where 36,935 of them from 333 object identities form the training set, and the test set consists of 18,290 bounding boxes from the other 333 identities.  The standard probe and gallery sets consist of 1,052 and 18,290 images respectively.  The synthetic dataset provided by AI City Challenge 2020 is generated by VehicleX, which is a publicly aviablable 3D engine. There are 192,150 images of 1362 vehicles are annotated with detailed labels in total. We select 50 vehicle identities of 3462 images in CityFlow as our self-val set. All the left images are used as the training set. So our vehicle Re-ID model self-training set has a total of 225,623 images of 1645 vehicle ids. Besides, after semi-supervised learning, the images in the test set are assigned 330 ids, finally we get a training set of 1975 ids.

\subsection{Implementation Details }
We totally trained several Re-ID models including DenseNet161 \cite{huang2017densely}, SE-ResNet152 \cite{hu2018squeeze} and  ResNet152 with MGN \cite{wang2018learning}. For some models, our training dataset is the officially provided training set and synthetic dataset. Due to the background redundancy of some vehicle images, we resize the image to 414$\times$310, and then randomly crop to 384$\times$288. At the same time, we perform random erasing to increase the difficulty of sample learning and horizontally flip the image to add diversity of the training images.

For other models, we use AI City Challenge 2019 first place Baidu's open-source refined dataset by their detector. In this part of the dataset, we resize the image to 384$\times$288, and no need to perform the operation of crop to background.

We adopt the strategy of random sample, each vehicle id selects  4 instance images in one batchsize. And we use Stochastic Gradient Descent (SGD) to train CNN  models in total of 1000 epochs. Strategies such as warm up learning rate and learning rate gradient attenuation are added to the training process. The initial learning rate is set to 0.03, which is decayed to 3$\times$10$^{-3}$, 3$\times$10$^{-4}$ and 3$\times$10$^{-5}$ at 300th epoch, 600th epoch and 900th epoch. We implement our model on PyTorch, and all the models are trained and tested on eight Tesla P40 GPUs.

\subsection{Semi-Supervised Learning,}
We use 225,623 images of 1645 vehicle ids train Re-ID model to extract the features of the test set. Given that there are 333 ids in the test set, k-means is used to cluster the test set images and 19,342 images of 333 vehicle ids are added to the original training set to retrain the Re-ID model. For each single model under different training sets, the mAP on the self-val set are shown in Table. \ref{table_1}. With the increase in the amount of training data, each of our single models has significantly improved on self-val set. However, because fake label is not very accurate, and the training model overfits the test set with certain incorrect labels, only DenseNet161 improves on the test set in the end, and as a single model of the last ensemble models. Model selection experiments are conducted in next section.

\vspace{6pt}

\makeatletter\def\@captype{table}\makeatother

\resizebox{0.44\textwidth}{!}{%
\begin{tabular}{|c|l|c|}
\hline
Methods & Train Dataset & \multicolumn{1}{l|}{self-val set mAP(\%)} \\ \hline 
\multicolumn{1}{|l|}{\multirow{2}{*}{DenseNet161}} & Train Set & 66.8 \\ \cline{2-3} 
 & Train Set + Test Set (fake label) & 72 \\ \hline
\multicolumn{1}{|l|}{\multirow{2}{*}{SE-ResNet152}} & Train Set & 71.2 \\ \cline{2-3} 
\multicolumn{1}{|l|}{} & Train Set + Test Set (fake label) & 72 \\ \hline
\multirow{2}{*}{ResNet152 with MGN} & Train Set & 66.7 \\ \cline{2-3} 
 & Train Set + Test Set (fake label) & 70 \\ \hline
\end{tabular}%
}
\vspace{4pt}
\caption{After adding the fake label test set to the training set, the comparison of mAP on the self-val set of each model.}
\label{table_1}

\begin{figure*}[htbp]
\centering
\includegraphics[width=\textwidth ]{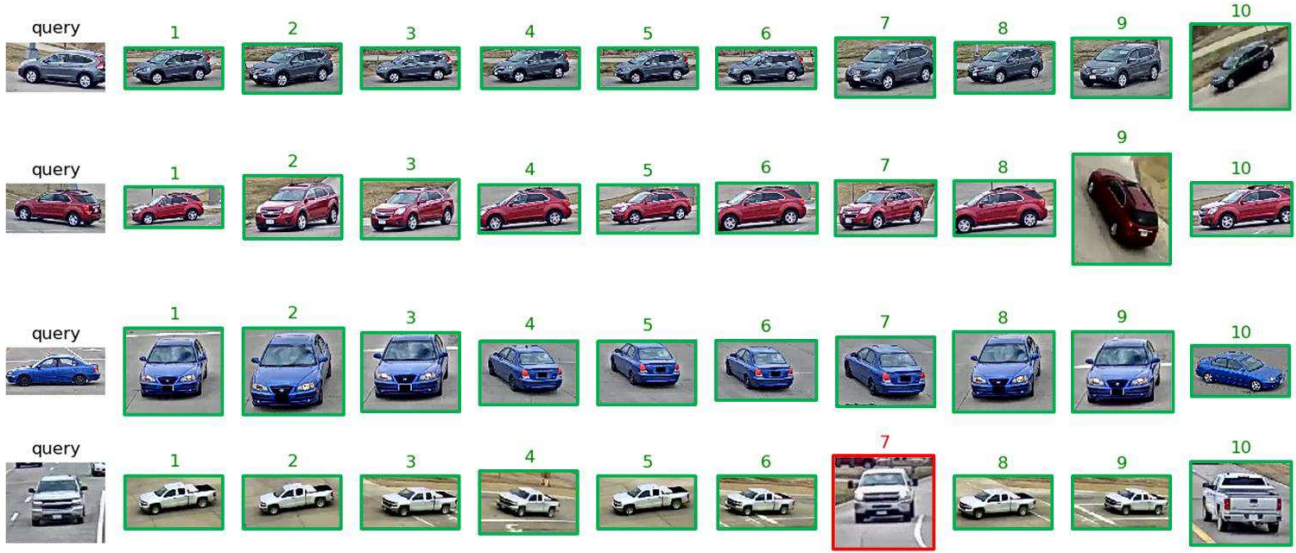}

\caption{Our vehicle Re-ID network rank the top-10 lists of four query images on the our self-val set of CityFlow dataset. The image with the green border boundaries belong to the same identity as the query, and the image with the red boundaries do not.}
\label{fig_3}
\end{figure*}

\subsection{Evaluation of Vehicle Re-Identification}

\textbf{Model Ensemble. }
We conduct experiments on different model ensemble strategies as shown in the Table. \ref{tabel_e}. \textbf{Average of feature merge}: take the average of the 2048-dimensional feature inferred from each single model. \textbf{Average of distance matrix}: take the average of the distance matrix calculated by the features of the query and gallery inferred from each single model. \textbf{Concatenated feature}: concatenate the 2048-d features output by each single model, expanding in dimension. Experiments show that Concatenated feature performs best.

Therefore, we concatenate the normalized features of SE-ResNet152 trained on the train set, ResNet152 with MGN (trained on the train set), DenseNet161 (trained on the train set) and DenseNet161 (trained on the train set with fake labe test set) for feature distance calculation, which is better than using a single model.

When the test set is labeled with fake labels, the clustering results are not refined, resulting in a large number of incorrect labels in the clustering results.

Although ResNet152 with MGN and SE-ResNet152 (trained on the train set with fake labe test set) perform better on self-val set than without fake label test set, but during training, the model overfits the test set with a large number of incorrect labels. Therefore, in the final model ensemble, the mAP accuracy decreases on the test set.

However, DenseNet161 does not  fit the test set with the incorrect labels, so we choose these 4 models listed in the Table. \ref{tabel_e} for the final ensemble feature.

\vspace{6pt}

\makeatletter\def\@captype{table}\makeatother
\resizebox{0.45\textwidth}{!}{%
\begin{tabular}{l|c|c}
\hline \hline
\multicolumn{1}{c|}{Methods} & Self-val set mAP(\%) & Test set mAP(\%) \\ \hline \hline
SE-ResNet152 (TT) & 71.2 & - \\ \hline
ResNet152 with MGN (TT) & 66.7 & - \\ \hline
DenseNet161 (TT) & 66.8 & - \\ \hline
DenseNet161 (TTF) & 72 & - \\ \hline \hline
Ensemble strategies &  &  \\ \hline \hline
Average of feature merge & 72.9 & - \\ \hline
Average of distance matrix & 74.2 & - \\ \hline
Concatenated feature & 74.8 & 66.84 \\ \hline \hline
\end{tabular}%
}
\vspace{4pt}
\caption{The mAP accuracy between single models and the ensemble model under different ensemble strategies on the validation set and test set. TT: trained on the train set, TTF: trained on the train set with fake labe test set. (In this table, we do not use re-ranking and query expansion on self-val set).}
\label{tabel_e}

\textbf{Post processing. }
We conduct the post-processing ablation experiment on the self-val set of CityFlow datasets, with SE-ResNet152 as the backbone of single model. It can be seen in Table. \ref{table_p} that each post-processing method has a certain improvement on  the  result. Although there is some improvement in query expansion on the self-val set, it is not reflected in the test set. Gallery feature merge will bring a certain improvement to the result in the test set. So in the end, the post-processing we do on the test set is shown in the testing stage of Fig. \ref{fig_1}.

\vspace{6pt}

\makeatletter\def\@captype{table}\makeatother
\resizebox{0.44\textwidth}{!}{%
\begin{tabular}{r|l|lllll}
\multicolumn{1}{l|}{} & \multicolumn{1}{c|}{SE-ResNet152} &  &  &  &  &  \\ \hline
Center Crop? &  & \checkmark & \checkmark & \checkmark & \checkmark & \checkmark \\
Image Flip? &  &  & \checkmark & \checkmark & \checkmark & \checkmark \\
Query Expansion? &  &  &  & \checkmark & \checkmark & \checkmark \\
Gallery Feature Merge? &  &  &  &  & \multicolumn{1}{c}{-} & \multicolumn{1}{c}{-} \\
Re-Ranking? &  &  &  &  &  & \checkmark \\ \hline
Self-val set mAP(\%) & \multicolumn{1}{c|}{71.1} & \multicolumn{1}{c}{71.6} & \multicolumn{1}{c}{72.4} & \multicolumn{1}{c}{74.2} & \multicolumn{1}{c}{-} & \multicolumn{1}{c}{83.8}
\end{tabular}%
}

\vspace{4pt}

\caption{Comparison of different post-processing methods on our self-val set of CityFlow datasets.}
\label{table_p}

\textbf{Impact of filter grafting. }
To verify the effectiveness of filter grafting on our vehicle Re-ID model, we have done some experiments on the SE-ResNet152. We try two models grafting, which will bring about 1\% improvement in mAP on the baseline. Due to the limitation of the time and GPU resources, no more grafting models are added and we did not try filter grafting on other backbones.

\subsection{Results of Vehicle Re-Identification}
\textbf{Compare with others on the AIC2020: }
Table. \ref{table_4} lists the ranks of our team and the results of our team on the vehicle Re-ID task of AI City Challenge 2020.

\vspace{6pt}

\makeatletter\def\@captype{table}\makeatother
\resizebox{0.44\textwidth}{!}{%
\begin{tabular}{|c|c|c|c|l}
\cline{1-4}
Rank & Team ID & Team Name & Score &  \\ \cline{1-4}
1 & 73 & Baidu-UTS & 0.8413 &  \\ \cline{1-4}
2 & 42 & ZOOZOO & 0.7810 &  \\ \cline{1-4}
3 & 39 & DMT & 0.7322 &  \\ \cline{1-4}
4 & 36 & IOSB-VeRi & 0.6899 &  \\ \cline{1-4}
\textbf{5} & \textbf{30} & \textbf{BestImage} & \textbf{0.6684} &  \\ \cline{1-4}
6 & 44 & BeBetter & 0.6683 &  \\ \cline{1-4}
7 & 72 & UMD$\_$RC & 0.6668 &  \\ \cline{1-4}
8 & 7 & Ainnovation & 0.6561 &  \\ \cline{1-4}
9 & 46 & NMB & 0.6206 &  \\ \cline{1-4}
10 & 81 & Shahe & 0.6191 &  \\ \cline{1-4}
\end{tabular}%
}

\vspace{4pt}

\caption{Ranking list of City-Scale Multi-Camera Vehicle Re-Identification in AIC2020.}
\label{table_4}

\textbf{Visualization of the results: }
The qualitative results on CityFlow our self-val set are shown in Fig. \ref{fig_3}. The top-10 ranking lists for our query images are visualized. From the results, we can see that our model successfully retrieves most of the vehicle pictures from the gallery.

\section{Conclusions}
We participated in the task of the NVIDIA AI City Challenge 2020: the vehicle re-identification task. Our solution for vehicle re-identification is based on training multiple CNNs with a few methods including multi-loss, filter grafting, part-based feature learning,etc. Then, we use semi-supervised method to annotate images in the test set with a fake label. Meanwhile, several post-processing methods are utilized to furthermake a progress of the result. Finally, we finished with 5-th place in the 2020 AI-City challenge for  City-Scale  Multi-Camera  Vehicle  Re-Identification.



\end{document}